\documentclass[preprint,3p,twocolumn,authoryear]{IEEEtran}
\usepackage{amssymb}  
\usepackage{hyperref}
\usepackage{ucs}
\usepackage{comment}
\usepackage{amsmath}
\usepackage{pbox}
\usepackage{tcolorbox}
\usepackage{bm}
\usepackage{graphicx}

\usepackage{array}
\usepackage{multirow}
\usepackage{colortbl}
\usepackage{tikz}
\usepackage{xspace}
\usepackage{soul}
\usepackage{diagbox}

\newcommand{\rl}{RL\xspace}

\newcommand{\imagine}{{\sc imagine}\xspace}
\newcommand{\decstr}{{\sc decstr}\xspace}
\newcommand{\gangstr}{{\sc gangstr}\xspace}
\newcommand{\hme}{{\sc hme}\xspace}

\newcommand{\zpd}{ZPD\xspace}

\newcommand{\irl}{Interactive RL\xspace}

\newcommand{\tom}{ToM\xspace}

\newcommand\pr[1]{\cellcolor{green!12} $\circ$}
\newcommand\ye[1]{\cellcolor{green!30} $\bullet$}
\newcommand\ot[1]{$\star$}
\newcommand\spa[1]{$\spadesuit$}
\newcommand\dn[1]{.}
\newcommand\no[1]{\cellcolor{red!30}{\bf x}}
\newcommand\bo[1]{$\bullet\star$}
\newcommand\na[1]{\cellcolor{black!25}{\sc N/A}}

\newcounter{cbox} 
\newcommand{\cbox}{\arabic{cbox}}

\newcounter{cmes} 
\newcommand{\cmes}{\arabic{cmes}}

\newcounter{cques} 
\newcommand{\cques}{\arabic{cques}}

\definecolor{myred}{rgb}{0.8,0,0}
\definecolor{mygreen}{rgb}{0,0.6,0}
\definecolor{myblue}{rgb}{0,0,0.7}

\definecolor{mynicered}{rgb}{0.8,0,0.2}

\usepackage{subfiles}
\usepackage{array, makecell}

\definecolor{deletecolor}{rgb}{0.5,0.5,0.5}

\begin{document}

\title{Towards Teachable Autotelic Agents}

\author{Olivier Sigaud, Ahmed Akakzia, Hugo Caselles-Dupr\'{e}, C\'{e}dric Colas, \\Pierre-Yves Oudeyer and Mohamed Chetouani
\thanks{Olivier Sigaud, Ahmed Akakzia, Hugo Caselles-Dupr\'{e} and Mohamed Chetouani are with Sorbonne Universit\'e, CNRS,
    Institut des Syst\`emes Intelligents et de Robotique, F-75005 Paris, France}
\thanks{C\'{e}dric Colas and Pierre-Yves Oudeyer are with
INRIA Bordeaux Sud-Ouest, \'equipe FLOWERS}
\thanks{Manuscript received XXX; revised YYY.}}

\maketitle

\markboth{IEEE TCDS, XXX}{Sigaud \MakeLowercase{\textit{et al.}}: Towards Teachable Autonomous Agents}

\begin{abstract}
Autonomous discovery and direct instruction are two distinct sources of learning in children but education sciences demonstrate that mixed approaches such as assisted discovery or guided play result in improved skill acquisition. In the field of Artificial Intelligence, these extremes respectively map to autonomous agents learning from their own signals and interactive learning agents fully taught by their teachers. In between should stand \textit{teachable autonomous agents} (\textsc{taa}): agents that learn from both internal and teaching signals to benefit from the higher efficiency of assisted discovery. Designing such agents will enable real-world non-expert users to orient the learning trajectories of agents towards their expectations. More fundamentally, this may also be a key step to build agents with human-level intelligence. This paper presents a roadmap towards the design of teachable autonomous agents. Building on developmental psychology and education sciences, we start by identifying key features enabling assisted discovery processes in child-tutor interactions. This leads to the production of a checklist of features that future \textsc{taa}s will need to demonstrate. The checklist allows us to precisely pinpoint the various limitations of current reinforcement learning agents and to identify the promising first steps towards \textsc{taa}s. It also shows the way forward by highlighting key research directions towards the design or autonomous agents that can be taught by ordinary people via natural pedagogy. 
\end{abstract}

\section*{Introduction}
\label{sec:intro}

From the etymology of the word, being \textit{autonomous} means deciding by oneself ({\em autos}) of its own rules ({\em nomos}). More generally, an agent can be said autonomous if it determines its own sensorimotor behavior, if it makes its own decisions.
Autonomy matters for Artificial Intelligence (AI). Indeed, at first glance, intelligence seems to require some autonomy: if we always had to tell an agent what to do at each step of a sequential decision or control process, we would not consider such an agent as intelligent. Let us temporarily consider a radical definition and call ``truly autonomous" an agent which would {\em only} decide what to do on its own, without any constraint, e.g. consideration for our needs and expectations, without ethics. This agent would be useless, perhaps even dangerous. At first glance, {\em true autonomy} and {\em usefulness} seem to be contradictory requirements: if an autonomous agent decides what to do only on its own, how can it be useful at all?

\begin{figure}[!ht]
  \centering
{\includegraphics[width=\columnwidth]{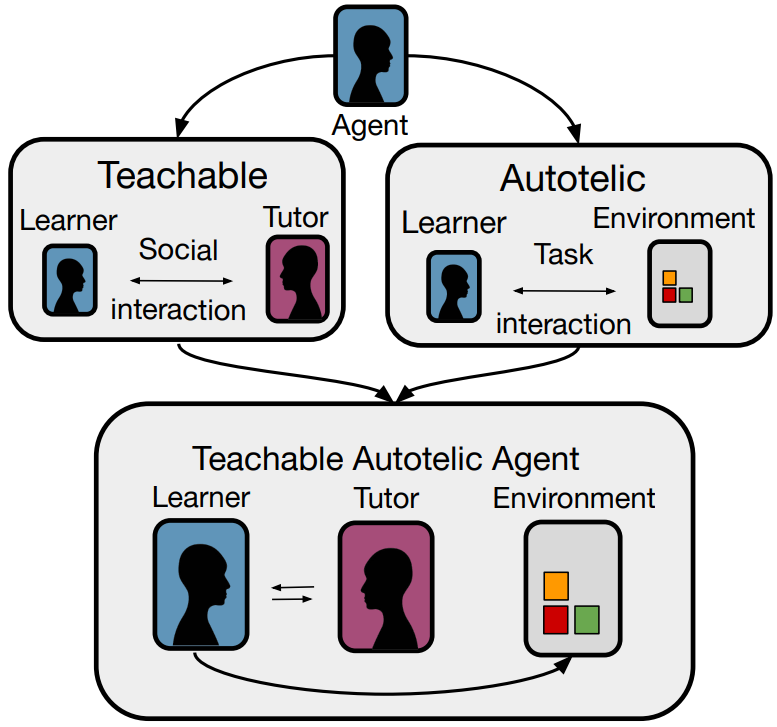}} \hfill
  \caption{Towards Teachable Autotelic Agents. We argue that teachability and autotelic learning are complementary components to reach human-level AI. To be fully teachable, agents must also be capable of inferential social learning.\label{fig:ttaa_init}}
\end{figure}

\begin{table*}[htp]
\centering
\caption{The teachability check-list. The table states whether the algorithms in the columns support the properties in the rows. \no{1} (red): no, \pr{1} (light green): preliminary, \ye{1} (green): yes, \na{1} (grey): not applicable. The column on autotelic agents considers isolated agents, thus none of the properties related to social interactions can be applied.}  \label{tab:teachability_checklist}
\resizebox{\textwidth}{!}{\begin{tabular}{|r|r|r|r|r|r|r|}
    \hline
    \backslashbox{Properties}{Agents} &\makecell{IRL\\ agents} & \makecell{Autotelic\\ Agents} & \makecell{Inferential\\ Social Agents} & \imagine & \decstr & \makecell{\hme +\\ \gangstr} \\ \hline
    \multicolumn{7}{c}{\textbf{Learner Properties (Autotelism)}} \\ \hline
    \multicolumn{1}{|l|}{Autotelic learning} & \cellcolor{black!25}\na{1} & \ye{1} & \pr{1} & \ye{1} & \ye{1} & \ye{1}   \\ \hline
    \multicolumn{1}{|l|}{Open-ended learning} & \no{1} & \ye{1} & \no{1} & \ye{1} & \ye{1} & \ye{1}   \\ \hline
    \multicolumn{1}{|l|}{Few shot learning} & \no{1}& \ye{1} & \no{1} & \ye{1} & \ye{1} & \ye{1} \\ \hline
    \multicolumn{1}{|l|}{Hierarchical learning} & \no{1} & \pr{1}  & \no{1} & \no{1} & \no{1} & \ye{1} \\ \hline
    
    \multicolumn{7}{c}{\textbf{Learner Properties (Social awareness)}} \\ \hline
    \multicolumn{1}{|l|}{Sensitivity to social signals} & \ye{1} & \na{1} & \ye{1} & \ye{1}  & \ye{1} & \ye{1} \\ \hline
    \multicolumn{1}{|l|}{Proficient language learning} & \pr{1} & \na{1} & \no{1} & \pr{1} & \pr{1} & \no{1} \\ \hline
    \multicolumn{1}{|l|}{Recognition of pedagogical signals} & \no{1} & \na{1}& \ye{1} & \no{1} & \no{1} & \no{1} \\ \hline
    \multicolumn{1}{|l|}{Observational learning} & \no{1} & \na{1} & \ye{1} & \no{1} & \no{1} & \no{1} \\ \hline
    
    \multicolumn{7}{c}{\textbf{Learner Properties (Social inference)}} \\ \hline
    \multicolumn{1}{|l|}{Modelling the tutor} & \no{1} & \na{1}  & \ye{1} & \no{1} & \no{1} & \ye{1} \\\hline
    \multicolumn{1}{|l|}{Internalization} & \no{1} & \na{1} & \no{1} & \no{1} & \no{1} & \ye{1} \\ \hline
    \multicolumn{1}{|l|}{Pragmatic learning} & \no{1} & \na{1} & \ye{1}& \no{1} & \no{1} & \no{1} \\ \hline
 
    \multicolumn{7}{c}{\textbf{Tutor Properties}} \\ \hline
    \multicolumn{1}{|l|}{Motivation Regulation} & \no{1} & \na{1}  & \ye{1} & \no{1} & \no{1} & \ye{1}  \\ \hline
    \multicolumn{1}{|l|}{\zpd management} & \no{1} & \na{1}  & \ye{1} & \pr{1} & \pr{1} & \ye{1} \\ \hline
    \multicolumn{1}{|l|}{Pedagogical demonstrations} & \no{1} & \na{1}  & \ye{1} & \no{1} & \no{1} & \no{1} \\ \hline
    \multicolumn{1}{|l|}{Modelling the learner} & \no{1} & \na{1}  & \ye{1} & \no{1} & \no{1} & \ye{1} \\ \hline
    
    \multicolumn{7}{c}{\textbf{Tutoring Process Properties}} \\ \hline
    \multicolumn{1}{|l|}{Social-based tutoring strategy} & \ye{1} & \na{1}  & \ye{1} & \pr{1} & \no{1} & \pr{1}  \\ \hline
    \multicolumn{1}{|l|}{Task-based tutoring strategy} & \ye{1} & \na{1}  & \ye{1}& \no{1} & \no{1} & \no{1} \\ \hline
     \multicolumn{1}{|l|}{Social communication-based transparency} & \ye{1} & \na{1}  & \ye{1} & \no{1} & \no{1} & \no{1}  \\ \hline
    \multicolumn{1}{|l|}{Task-based transparency} & \ye{1} & \na{1}  & \ye{1} & \ye{1} & \no{1} & \ye{1} \\ \hline
\end{tabular}}
\end{table*}

This apparently rhetorical question can be turned into a much more practical one: if a truly autonomous agent decides on its own, how can we influence it to make it useful anyways? Charkraboti et al. proposed that autonomous agents should understand and adapt to human behavior much like humans adapt to the behavior of other humans \cite{chakraborti2017ai}. How to obtain such an adaptation?
Part of the answer can be found in the conclusion of the seminal paper of Alan Turing about Artificial Intelligence:
\begin{quote}
   ``{\em It can also be maintained that it is best to provide the machine with the best sense organs that money can buy, and then teach it to understand and speak English. That process could follow the normal teaching of a child.}'' \cite{turing1950computing} 
\end{quote}

Even if they are not always autonomous in the common sense\;---\;they may need caregivers to fulfill their basic living requirements\;---\;children are definitely autonomous in the sense that we cannot fully control their behaviour. Nevertheless, we can succeed in influencing their behavior through many ways, including ``normal teaching.'' 

But again, we cannot teach an agent if it is ``radically autonomous" in the sense outlined above. So, resolving the contradiction between {\it true autonomy} and {\it usefulness} requires a weaker notion of autonomy. In the following sections, we propose to equate \textit{autonomy} with the concept of \textit{autotelicity}: the ability to set one's own goals and to learn to achieve them using one's own learning signals \cite{steels2004autotelic,colas2020intrinsically}. Autotelic agents are equipped with forms of intrinsic motivations enabling them to represent, generate and pursue their own goals.
Though autotelic agents pursue their own goals, their behavior can still be influenced by external signals. Such agents could be made appropriate and useful through teaching if we can influence its goal representations or goal sampling strategy.
Thus this paper proposes to reach useful autonomous agents by developing \textit{teachable autotelic agents}\;---\;agents that choose their own goals but whose choice still benefits from teaching via natural social interactions, see \figurename~\ref{fig:ttaa_init}.

According to education sciences, children can learn in three ways: \textit{direct instruction} where a tutor explicitly sets the learning goals of children step by step,
\textit{unassisted discovery}\;---\;also called \textit{free play}\;---\;where children are left on their own to discover new things, and \textit{guided play} or \textit{assisted discovery}, where the tutor intervenes on the discovery process of children to make it more fruitful \cite{weisberg2016guided, yu2018theoretical}.


In this paper we take the stance that, when transposed to AI research, the first way children learn under the strict guidance of a caregiver or tutor corresponds to interactive learning, and particularly interactive reinforcement learning research \cite{lin2020review}, whereas the second way, learning on their own, corresponds to research on autotelic reinforcement learning agents \cite{colas2020intrinsically}. However, the third way, guided play, has no counterpart in current AI research though it is believed to be the most efficient \cite{yu2018theoretical}. A major step towards endowing artificial agents with such capabilities consists in combining the properties of interactive and autotelic reinforcement learning agents. We will describe preliminary efforts in this direction. However, this integration remains insufficient. One must also keep in mind that children are inferential social learners and that social inference mechanisms play a key role in the extraordinary learning capabilities of children \cite{gweon2021inferential}.

In practice, AI agents will need such capabilities. When immersed in human societies, they will need to acquire the various socio-cultural skills required in these ecosystems\;---\;skills specific to regions or social groups. The only way to do so is by learning them through practical interactions with social partners. For these agents, being teachable, autonomous and capable of social inferences means being equipped with the core capabilities to acquire such skills.

Beyond this, a deeper, more fundamental reason for endowing AI agents with such capabilities stems from the endeavour of building a human-level AI \cite{lake2017building}. It might be the case, as put forward by the {\em social situatedness} vision of researchers like Vygotsky \cite{vygotsky1978mind}, Bruner \cite{bruner1990acts, bruner1991narrative, bruner2009process}, Tomasello \cite{tomasello2009constructing} and others \cite{dautenhahn1995getting, zlatev2001epigenesis} that, in addition to the capability to pursue their own goals, social interactions with caregivers, tutors and mates are themselves necessary conditions for the emergence of sophisticated forms of intelligence in agents, see \cite{lindblom2003social} for a review. In particular, these social and cultural interactions may play a crucial role in the acquisition of shared cognitive representations making sense for agents and their human partners. From that perspective, to obtain useful autonomous agents in a strong sense, we should focus on questions centered on the role of social interactions in the acquisition of representational capabilities compatible with those of social partners \cite{vygotsky1978mind,tomasello2009constructing}.

Our roadmap towards the design of such agents is organized as follows. In Section~\ref{sec:teach}, we build on the developmental psychology and education sciences literature to extract a \textit{teachability checklist} emerging from the natural teaching of children. More precisely, we organize these properties in three categories according to whether they involve the learning child, the tutor or the whole tutoring process. We argue that these properties are needed to teach autotelic agents. 
In Section~\ref{sec:or}, we scrutinize current research in Interactive Reinforcement Learning, Autotelic Reinforcement Learning and Inferential Social Learning under the light of these identified properties.
We then outline in Section~\ref{sec:and} the emergence of a new line of research striving to combine the properties of interactive agents and autotelic agents. In particular, we describe the \imagine \cite{colas2020language}, \decstr \cite{akakzia2021grounding} and \gangstr \cite{akakzia2022help} agents as well as the the \hme protocol \cite{akakzia2022help} as examples of autotelic agent research striving to endow AI agents with both problem solving capabilities and primitive interactive learning capabilities. 
The main outcome of this work is Table~\ref{tab:teachability_checklist}, where we recap all the  surveyed approaches and state whether they demonstrate the properties expected from teachable autonomous agents. In Section~\ref{sec:discu}, we build on the table and highlight key research directions towards the design or autonomous agents that can be taught by ordinary people via natural pedagogy.


\section{The ``Normal Teaching of a Child"}
\label{sec:teach}

This section leverages observations from developmental psychology to characterize how children learn and how tutors contribute to their learning process. First, we extract the properties of learning children (Section~\ref{sec:child_iima}), then those of the tutors (Section~\ref{sec:prop_tutor}) and finally, those of tutoring interactions themselves (Section~\ref{sec:prop_process}).

\subsection{Properties of Children Learners}
\label{sec:child_iima}

In the list of properties below, the first four are related to the task learning capabilities of children and the rest to their social learning capabilities. 


\vspace{3mm}
\par\par\noindent {\bf Children are autotelic learners:}
Either through free or assisted play, children engage in sensorimotor interactions with their environment and discover new things which they might later try to reproduce.
During free-play, children rely on their intrinsic motivations to spontaneously explore their surroundings and unlock new reachable goals \cite{berlyne1966curiosity,gopnik1999scientist,chu2020play}. They automatically take ownership of these discovered goals, try to build their own understanding of them and attempt to pursue them. This process is very important for subsequent learning under the guidance of a tutor. For instance, the experiments in \cite{wood1976role} start with a short period of free-play with the experimental setup. This time is necessary for them to build an understanding of their surrounding before they can be influenced by social signals.

\vspace{3mm}
\par\noindent {\bf Children are open-ended learners:}
An extraordinary property of natural learning in children is that it is {\bf open-ended}: the child can solve new problems of increasing difficulty up to becoming an adult and keeps
learning during their whole life. 
Given the potentially infinite set of goals that they may pursue, children have to select some goal at all times. For that, they may attribute to potential goals a value of interest that evolves with time, resulting in efficiently organizing their own \textit{developmental learning trajectory} \cite{piaget1977development,thelen1996dynamic,smith2005development}. In other words, they self-define a learning curriculum that makes them very sample efficient: they avoid spending too much time on goals that are either too easy or too difficult, focusing on goals that present the right level of complexity at the right time.

\vspace{3mm}
\par\par\noindent {\bf Children are few shot learners:}
Children can transfer what they have learned from solving one task to a novel one. This flexibility allows them to benefit from some prior knowledge when facing a new problem that looks like the one they already know. Consequently, children are {\bf few shot learners}: they can leverage what they learned in previous tasks to master new tasks in a few trials. Thus, these mechanisms allow children to discover highly complex skills such as biped locomotion, block stacking or tool use, which would have been extremely difficult to learn if they had directly addressed these goals before mastering simpler skills.

\vspace{3mm}
\par\noindent {\bf Children are hierarchical learners:}
It is often the case that our tasks in everyday life have a hierarchical structure. For instance, the block assembly task of \cite{wood1976role} involves several repetitions of the same basic block manipulation movements.
More generally, the idea that children are {\bf hierarchical learners} is pervasive in the developmental psychology literature \cite{eppe2022intelligent}. The elementary skills mastered by children are often stepping stones for discovering how to learn other skills of increasing complexity. As Bruner writes:
\begin{quote}
    ``{\em The acquisition of skill in the human child can be fruitfully conceived as a hierarchical program in which component skills are combined into `higher skills' by appropriate orchestration to meet new, more complex task requirements.}'' \cite{bruner1973organization}
\end{quote}

\vspace{3mm}
\par\par\noindent {\bf Children are social learners:}
Even shortly after birth, newborn infants can imitate complex facial expressions such as happiness, sadness and surprise \cite{field1983discrimination, meltzoff1983newborn, meltzoff1988infant}. Infants can detect caregiver's eyes and prefer to look at pictures of direct gaze over averted gaze \cite{Farroni9602}. The developmental psychology literature reported several evidence of children's social responsiveness during interaction with caregivers \cite{bruner1973organization, vygotsky1978mind, tomasello2009constructing}. This literature demonstrates children's {\bf social sensitivity}, in particular when actions are directed to them such as for infant-directed speech \cite{saintgeorges2013}.

\vspace{3mm}
\par\noindent {\bf Children are proficient language learners:}
Only humans master complex compositional and recursive language. Children's puzzling ability to learn it so rapidly seems to be essential to their development. Indeed, children born deaf with no access to recursive sign language and Romanian children socially abandoned in Ceausescu’s orphanages showed decreased abilities for abstract compositional thinking and mental simulation \cite{vyshedskiy2019language}. The mastery of compositional language seems to further support other cognitive functions such as creativity \cite{chomsky_syntactic_1957,hoffmann_creativity_2018} or analogical reasoning \cite{gentner_analogy_2017}. 

\vspace{3mm}
\par\noindent {\bf Children recognize pedagogical signals:}
Children are able to recognize the pedagogical stance of a tutor from the general context of the initiated interaction and from communicative signals of the action itself \cite{csibra2009natural}. Children are sensitive to ostensive cues such as gaze, pointing or gesture modulation (``{\em motionese}" \cite{nagai2007can}), which are generated by the caregivers to signal their pedagogical intent. This specific strategy of caregivers and the children sensitivity to it is called {\bf natural pedagogy} \cite{csibra2009natural}. The children is able to infer the communicative intention and correctly interpret the instrumental intention to learn from the pedagogical caregiver. 

\vspace{3mm}
\par\noindent {\bf Children are observational learners:}
In natural interactions, children observe others acting in the environment even when no pedagogical intention is present, and still extract a lot of information from these observations, a process referred to as {\bf observational learning} \cite{varni1979analysis, meltzoff1999born}.

\vspace{3mm}
\par\noindent {\bf Children construct a model of the tutor:}
During assisted-play, it is crucial that children adjust their understanding of the task at hand using the available social signals. As \cite{wood1976role} put it, ``{\em children understand goals before being able to produce them}.''

\begin{figure}[!ht]
  \centering
{\includegraphics[width=\columnwidth]{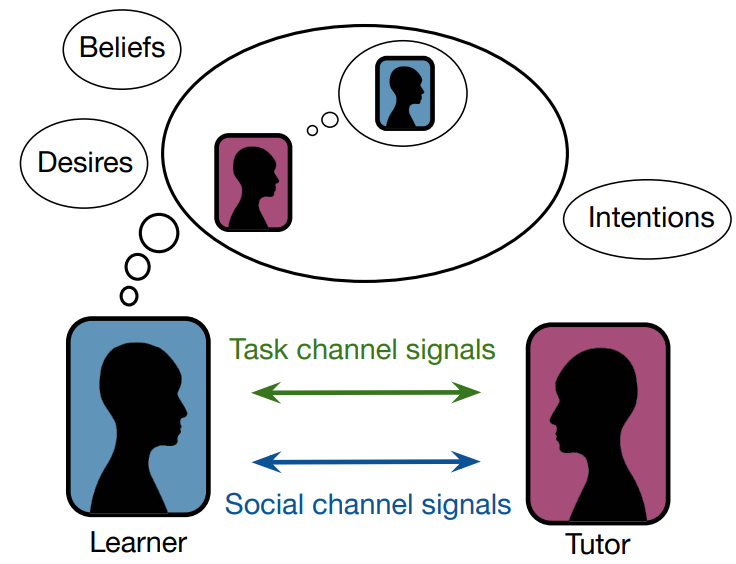}} \hfill
  \caption{Learners build and maintain a model of tutor's beliefs about their own knowledge, capabilities, intentions and desires. Tutors also have a model of learners, and maintain it to provide adapted teaching feedback.\label{fig:theory-of-mind}}
\end{figure}

Understanding the goal from the behavior of a tutor mostly relies on inferring the tutor's expectations and reasoning to figure out how to meet them. Thus, this process clearly calls upon a mental model of the tutor's expectation.
Such a mental model derives from a more general model called a {\bf Theory of Mind} (\tom), as illustrated in \figurename~\ref{fig:theory-of-mind} and put forward in recent developmental psychology papers \cite{velez2021learning, gweon2021inferential}. Having a \tom means being capable of reasoning about other people’s mental states \cite{jara2019theory}. The most common mental states these theories refer to are beliefs, desires and intentions.

\vspace{3mm}
\par\noindent {\bf Children internalize social signals:}
In Vygotsky's theory of development, higher-level cognitive capacities first appear as interpersonal processes before children internalize them and turn them into \textit{intra-personal psychological tools} \cite{vygotsky1978mind}. Parents first narrate the child's activities, orient their attention, keep them motivated, or decompose tasks for them. As children grow, they progressively internalize this social narration into private speech (outer speech for oneself) and, eventually, inner speech. Just like we use traditional tools to augment our control on the physical world, psychological tools augment our control on our own thoughts and behaviors. In a wealth of studies, private speech was proved instrumental to children's ability to reason and solve tasks. They use it for planning \cite{vygotsky1978mind, sokolov2012inner}, and even more so when the task gets harder \cite{berk_why_1994}. Children seem to internalize models of their tutors and self-generate tutor-like guidance and judgements on their own behavior.

\vspace{3mm}
\par\noindent {\bf Children are pragmatic learners:}
Several striking experiments have shown that infants use probabilistic inference guided by an intuitive understanding of how other people think, plan and act \cite{gweon2021inferential}. When presented with teaching signals, they consider how the information is generated, by whom, for whom, and why \cite{gweon2010infants}. Children also leverage a causal understanding of how and why those behaviors came to be, that is, a generative model of other minds. Children consider tutor's mental states (i.e., goals, beliefs, desires) but also their utilities (i.e., costs and rewards). Given this evidence, they are able to efficiently infer the intent that is communicated to them, and thus rapidly understand the task at hand \cite{gweon2021inferential}. All these properties make them {\bf pragmatic learners} and play a key role in their learning efficiency.

\subsection{Properties of Tutors}
\label{sec:prop_tutor}

Up to now, we have focused on the properties of children as efficient learners. We now investigate the properties of tutors.
We do not aim to design an artificial tutoring agent, but to extract from this perspective the properties that help a teachable agent better respond to natural tutoring signals, coming either from a human or another artificial agent.

\vspace{3mm}
\par\noindent {\bf Tutors regulate children's motivation:}
\cite{wood1976role} outline that most tutoring interactions are targeting {\bf motivation regulation} in children. They intend to keep them engaged in pursuing the instructed goal rather than their own goals. In turn, children must have developed their own interests beforehand, independently from social pressure. 
To illustrate this, \cite{sobel2010importance} oppose a \textit{discovery condition}, where free-play precedes social interaction, to a \textit{confirmation condition}, where it is the opposite: the tutor shows what to do and then leaves children on their own. They show that freely acting on a toy beforehand allows children to construct their own motivations, which then can be further regulated if needed in guided play. \cite{wood1976role} also account for the discovery condition when children are left to freely engage and get familiarized with the task before the tutoring process kicks off.
As the latter starts, the authors identify three processes that regulate the children's motivations. Through {\em recruitment}, the tutor should find a way so that the child engages into the targeted goal rather than its other own goals. Through {\em direction maintenance}, the tutor ensures that the child keeps committing to that specific goal, providing incentives to make further progress towards the goal. Finally, {\em frustration control} is meant to prevent children from giving up on the instructed goal.

\vspace{3mm}
\par\noindent {\bf Tutors maintain children in their zone of proximal development:}
In the construction domain, a scaffolding is a temporary physical structure used to support a working crew. The term has been borrowed by education researchers to refer to assistance provided by a tutor to support learning \cite{wood1976role, vygotsky1978mind, winn1994promises, benson1997scaffolding}. The two key aspects of scaffolding are 1)~helping the learner with yet unmanageable skills while 2)~allowing them to do as much as possible without help. First, in order to grasp initially hard tasks, the tutor can, for example, break them into more manageable sub-parts or engage in a \textit{thinking aloud} process, providing guidelines on how the task should be performed \cite{rosenshine1992use}. Second, by leaving them unassisted as much as possible, the tutor helps learners take responsibility over the task. This makes scaffolding a \textit{temporary} process, eventually enabling the learner to work independently.

Embedded within the scaffolding process is Vygostsky's concept of the \textit{zone of proximal development} (\zpd) \cite{vygotsky1978mind}. The \zpd is defined as the area between what the learner can accomplish on its own and what can be accomplished with the help of a tutor. Notably, the \zpd is always shifting as the child learns. As a result, tutor interventions must constantly be individualized to address this change and ensure {\bf \zpd management} until children eventually internalize the information and get exclusively self-regulated.

\vspace{3mm}
\par\noindent {\bf Tutors are modelling the tutee:}
To efficiently regulate the motivational system of children and provide appropriate instructions bringing them towards their \zpd, the tutor needs to monitor a model of the knowledge, hypotheses and performance of children as well as a model of the task itself: ``{\em The effective tutor must have at least two theoretical models to which he must attend. One is a theory of the task or problem and how it may be completed. The other is a theory of the performance characteristics of his tutee.}'' \cite{wood1976role}.
{\bf Modelling the learner} helps the tutor interpret what children are trying to do, so as to efficiently teach them. The authors go further and consider that this interpretation process consists in generating hypotheses about the behavior of children, something supposedly intuitive for humans.

\vspace{3mm}
\par\noindent {\bf Tutors are pedagogical:}
\cite{wood1976role} provide several cues showing that natural tutoring interactions do not rely much on demonstrations to be followed blindly. They observe that, among the 30 children of their study, ``{\em there was not a single instance of what might be called blind matching behaviour}.'' Blind matching behavior is what would be observed if children where replaying the tutor's trajectory without understanding the goal. Second, the authors mention that the only acts that children imitate are those they can already perform fairly well. That is, imitating is not a way to learn how to perform the tutor's actions, it is more a way to move to the next problem solving step.
Beyond showing what to do next, demonstrations also play a role in learning new skills. If they are not used for blind imitation, how do they help? ``{\em Demonstrating or `modelling' solutions to a task, [...] involves an `idealization' of the act to be performed and it may involve completion or even explication of a solution already partially executed by the tutee himself}'' \cite{wood1976role}.
In fact, the tutor is `imitating' in idealized form an attempted solution tried (or assumed to be have been tried) by children in the expectation that they will then `imitate' it back in a more appropriate form. That is, the tutor builds on a model of the learner's knowledge to communicate on the problem solving steps that are still inadequate. In other terms, the tutor shows rather than it does, which characterizes {\bf pedagogical teaching}.

\subsection{Properties of the Tutoring Process}
\label{sec:prop_process}

Finally, we extract properties of the tutoring process as interactions between tutors and their tutees, building on the social learning perspective of \cite{bandura1977social}. Tutoring can be conceptualized as a mutual exchange process using two main communication channels, the \textit{social channel} and the \textit{task channel}, were both partners can play alternatively the role of the emitter or the recipient, see \figurename~\ref{fig:theory-of-mind}.

\vspace{3mm}
\par
\begin{figure}[!ht]
  \centering
{\includegraphics[width=\columnwidth]{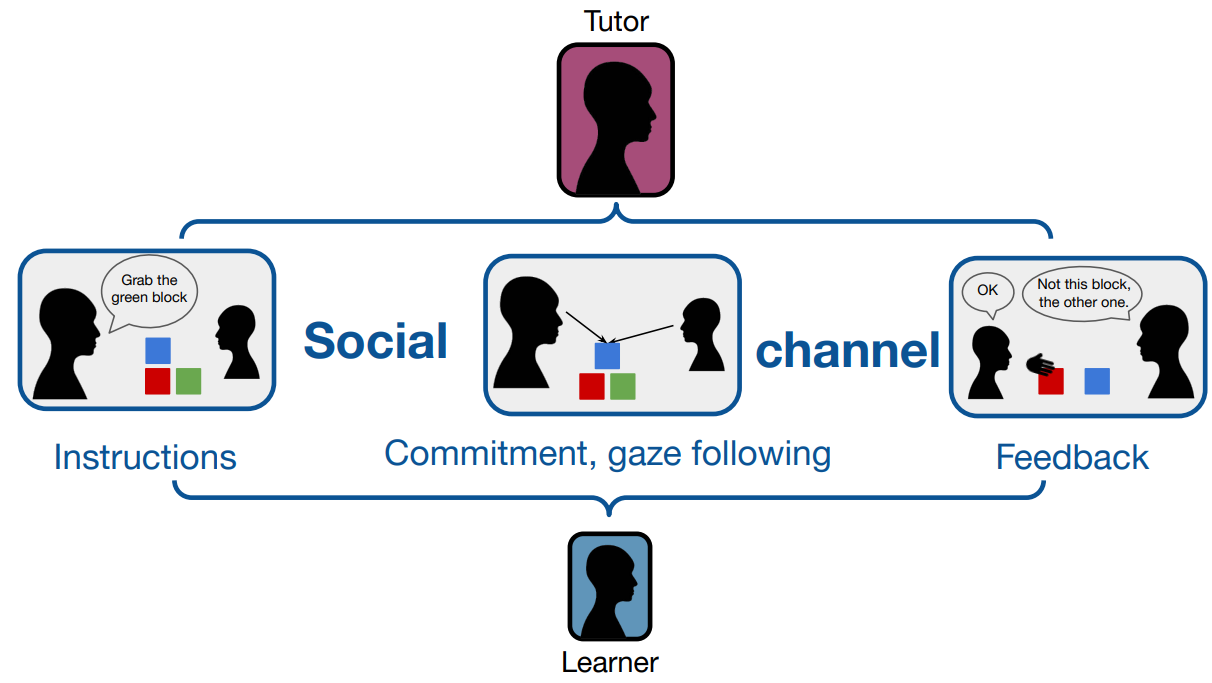}} \hfill
   \caption{Examples of social channel signals exchanged between a tutor and a learner. The signals can be verbal, such as instructions or feedback, or non-verbal, such as gaze following. These interaction signals are used to maintain the learner engaged into the learning activity. \label{fig:social-channel-signals}}
\end{figure}

\begin{figure}[!ht]
  \centering
{\includegraphics[width=\columnwidth]{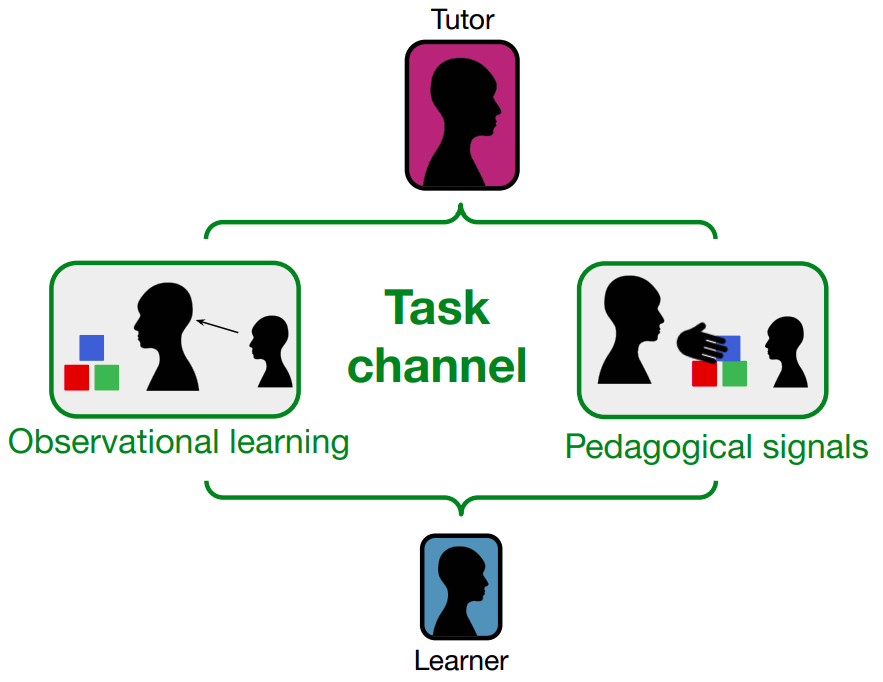}} \hfill
   \caption{Examples of task channel signals exchanged between a tutor and a learner. Here, these signals consist of the task-related behavior of the tutor, either as demonstrations or as performance without a pedagogical intention.\label{fig:task-channel-signals}}
\end{figure}

\noindent {\bf Tutoring combines the social and task channels:}
To be successful, tutoring should exploit and combine both the social and the task channels. On one hand, the social channel involves instructions, feedback or gaze (\figurename~\ref{fig:social-channel-signals}). It allows learners to not only adjust their own understanding of the task through non-motor signals, but also constructively learn to understand the tutor’s signals. On the other hand, the task channel involves watching motor interactions with the environment performed with the intent to teach or not (\figurename~\ref{fig:task-channel-signals}). Besides, children learn to adjust their beliefs from goal-related signals, thus fostering their proactive communication abilities. 

\vspace{3mm}
\par\noindent {\bf Partners are emitters and recipients:}
Both participants usually alternate the emitter and recipient roles, resulting in a mutual exchange (\figurename~\ref{fig:theory-of-mind}). For example, when learners execute a task in order to obtain a feedback from the tutor, they are first the provider and then the recipient. In more details, the tutor may emit through the social channel when providing instructions or non-verbal feedback, and through the task channel when performing or showing the task. We call the former {\bf social-based tutoring strategy} and the latter {\bf task-based tutoring strategy}. From the other side, the learner emits through the social channel when providing feedback on their understanding or asking questions. We call this {\bf  social communication-based transparency}. The learner emits through the task channel when imitating the tutor or performing the task to display their current capabilities, we call it {\bf task-based transparency}. Note that when both the tutor and the learner are artificial agents, since the learner has to perform the task, the presence of task-based transparency corresponds to the fact that the tutor reacts to the behavior of the learner.
Combining both channels results in various ``frames" of exchanges where both participants can use both channels and both roles. These frames of exchanges being goal-oriented, they are called ``{\em pragmatic frames}" in \cite{vollmer2016pragmatic}.


\section{Interactive, Autotelic and Inferential Social Agents}
\label{sec:or}

After this overview of natural tutoring processes, we turn to the AI research dedicated to the design of artificial learners and investigate for three broad classes corresponding to the first columns of Table~\ref{tab:teachability_checklist} how much the account for the properties we have listed.

\subsection{Reinforcement learners}

Reinforcement learning (\rl) is a process by which an agent learns to solve sequential decision problems from a reward signal \cite{sutton1998introduction}. The learning agent is initially ignorant of the consequences of its actions and must explore its environment to discover them. By maximizing future rewards, it progressively learns to favor more rewarding actions while avoiding costly ones.
RL first appeared as a computational model of learning by trial-and-error in rodents \cite{thorndike1911animal}, but is also useful to explain conditioning phenomena in monkeys \cite{schultz1997neural} and human decision making \cite{daw2006computational}. Thus, \rl seems to be a natural framework for modelling learning to solve problems in children. However, this framework suffers from several limitations. 

First, the behavior of \rl agents is fully determined by the reward. In the standard framework, this reward is externally provided by a human designer with some specific goal in mind. In the absence of this external reward, most \rl agents would learn nothing. As outlined in Section~\ref{sec:child_iima}, this is to be contrasted with children who can set their own goals and learn in full autonomy.

Second, \rl agents optimize a predetermined reward function. Their behavior should converge to a corresponding optimum and stop changing, which contrasts with the open-ended learning capabilities of children. At first glance, the richer framework of multitask \rl \cite{caruana97multitask} and its derivatives such as meta-\rl \cite{finn2018learning} seem to do better but, as long as the set of tasks is bounded, the obtained behavior should also converge to a steady optimum. To overcome these limitations, open-ended learning approaches suggest that \rl agents receive a potentially infinite sequence of unknown tasks \cite{doncieux2018open}. However, this framework does not explicitly answer one of the most important questions: where should all these tasks and reward functions come from? Practically, all existing approaches revert to the bounded learning case: first define a (bounded) space of reward functions, then sample from it. Indeed, it is hard to imagine and define a space of reward functions that would encompass all activities humans are able to pursue. Even in that case, how should we sample appropriate tasks for an agent along its lifetime? In contrast, humans seem to generate their own goals, learn from their own reward signals and organizing their own learning trajectories in a virtually infinite space of tasks.

Third, \rl is notoriously slow and sample inefficient \cite{botvinick2019reinforcement}. The groundbreaking successes that have made the field popular these last years have all been obtained with weeks of heavily parallel computations that would correspond to centuries of human experience. This is to be contrasted with the few shot learning capabilities of animals \cite{gruber2019new} and particularly humans \cite{csibra2009natural}.

Finally, the standard \rl framework accounts for an agent learning in isolation. As a consequence, standard \rl agents lack all the social properties linked to tutoring interactions that we outlined in Sections~\ref{sec:prop_tutor} and \ref{sec:prop_process}.

Now that we have established the limitations of the standard \rl framework, let us describe three AI research lines tackling them:  interactive reinforcement learning, autotelic learning and inferential social learning.

\subsection{Interactive Reinforcement Learners}
\label{sec:irl}

Agents immersed in the real-world cannot be preprogrammed to meet all their users' expectations. Interactive learning research tackles that fundamental problem by enabling non-expert users to communicate or teach their preferences and expectations in a natural way, as they would do with children \cite{vollmer2016pragmatic}. Thus, the field of interactive learning investigates and models the way a human tutor guides the learning process of an agent by providing teaching signals \cite{breazeal2008learning}. More specifically, \irl research focuses on the case where the agent is an \rl agent. 

From the perspective adopted in this paper, \irl can be seen as solving several of the issues outlined above.
First of all, by definition, \irl agents are {\bf social learners}. Though not all \irl agents are endowed with all these capabilities, one can find works where \irl agents benefit from {\bf social-based tutoring strategy} \cite{knox2010combining, grizou2014interactive, najar2016training} or {\bf task-based tutoring strategy} \cite{abbeel2004apprenticeship, argall2009survey} and works in which the \irl agent displays some {\bf task-based transparency} \cite{petit2012coordinating,wallkotter2021explainable} and {\bf social communication-based transparency} \cite{boucher2012reach, broekens2019towards,wallkotter2021explainable}.

The \irl approach relies too much on the tutor to drive the learning process and fails to account for the {\bf autonomy} of children. As for their {\bf open-ended} learning properties, one may consider that a human user may specify a sequence of increasingly more difficult tasks, for instance by specifying preferences about outcomes \cite{christiano2017deep}. 
But in the absence of autonomous learning, it is too much load for the tutor to always monitor the learning progress of an agent and to provide new tasks along a potentially infinite learning trajectory. Thus, \irl accounts better map to the {\em direct instruction} approach to education than to assisted discovery.

Along the same line, though teaching signals can substantially accelerate learning, \irl research does not generally consider multitask nor hierarchical contexts. Thus the corresponding agents do not display {\bf few shot learning} nor {\bf hierarchical learning} capabilities.

In addition, \irl research does not satisfactorily account for the social interaction properties of the tutoring processes outlined in Section~\ref{sec:prop_process}.

First, a well-known weakness of \irl research is that the way naive human users tend to teach an agent is far from meeting the expectations of the \rl framework \cite{thomaz2008experiments, thomaz2008teachable}.

Second, the way \irl research accounts for task-based tutoring strategy is called ``Learning from Demonstration'' (LfD) \cite{argall2009survey}. In this approach, an expert first performs highly rewarded trajectories in the environment where the task is defined. Then, data from these trajectories are collected and fed into the replay buffer of the learning agent\;---\;a sort of episodic memory. From this data, the agent learns an efficient policy with RL as if it was its own memories \cite{hester2017deep, vecerik2017leveraging}. Rather than using RL, an alternative approach called ``Behavioral Cloning'' (BC) consists in \textit{cloning} the imitated policy by applying standard \;---\;another kind of machine learning process\;---\;from the same data to directly obtain a policy which behaves like the imitated one \cite{torabi2018behavioral}. These methods assume the experience of the expert to be directly transferred into the memory of the learning agent, which cannot happen yet in real life given that both participants have different viewpoints. Besides, they assume that the tutor and the learner share the same state space, action repertoire and dynamics of interaction with the environment. This is unlikely in real life situations as made obvious by works on the so called ``correspondence problem" \cite{nehaniv2002correspondence}. Finally, they assume that the agent can perfectly observe the states and actions of the demonstrator and imitate these actions, in sharp contrast with what we outlined in Section~\ref{sec:prop_tutor}. It is more likely that natural learners recognize the goals of the partner and try on their own to reach these goals. This is known as {\em goal emulation} \cite{tomasello1998emulation, ugur2011goal} and can be accounted for in the \rl framework through inverse \rl processes \cite{abbeel2004apprenticeship}. Among other things, this approach can help solving the correspondence problem.

Beyond this, in \irl, teaching signals lack a communicative intent. As outlined in Section~\ref{sec:teach}, infants and more generally humans are sensitive to pedagogical teaching, where the teaching signals are specifically emitted to optimize learning efficiency. This is overlooked in the \irl literature, where a tutor will provide the same demonstrations to all learners without taking their current knowledge in consideration, thus failing to account for the {\bf \zpd management}, {\bf pedagogical teaching}  and {\bf modelling the learner} properties. Reciprocally, these methods expect all learners to learn equally well from a given set of demonstrations, thus \irl agents cannot be seen as {\bf pragmatic learners}.

Finally, natural teaching methods build on linguistic description of behaviors, instructions, explanations and both verbal and nonverbal feedback. As for the linguistic signals, \irl agents rely on a limited set of predefined tokens, which cannot be confused with a form of {\bf language proficiency}.
For educating agents with the richer signals used in natural teaching, these agents must be equipped with richer capabilities. 
We claim that having the capability to represent and pursue goals, to autonomously imagine and select goals, to infer the goals of others and to interact with them about these goals are some of the required capabilities, as these goals can play a pivotal role at the interface between user expectations and autonomous behavior learning. These concerns play a key role in the emergence of the line of research that we describe in Section~\ref{sec:and}.

\subsection{Autotelic Reinforcement Learners}
\label{sec:pb_solvers}

The extraordinary transition from the mental life of human infants to the sophisticated intelligence of adults is mostly modelled in the domain of developmental robotics and AI \cite{weng2001autonomous, zlatev2001epigenesis, lungarella2003developmental}.
A central line of research in this domain is interested in the design of autotelic agents \cite{steels2004autotelic, schembri2007evolving}. These embodied agents interact with their environment at the sensorimotor level and are provided with the ability to represent and set their own goals and rewarding themselves when they achieve them \cite{oudeyer2007intrinsic, forestier2017intrinsically,colas2020intrinsically}.
By definition, they are {\bf autotelic learners}.

Fundamentally, these agents are problem solvers. Implementing their learning capabilities using \rl is natural, since the \rl framework provides the model of choice to account for problem solving capabilities \cite{sutton1998introduction}. 
Most of these agents are equipped with one or several goal spaces and rely on goal-conditioned \rl \cite{colas2020intrinsically} and automatic curriculum learning \cite{portelas2020automatic} to learn to achieve those goals along an open-ended developmental trajectory. This endows them with the capability to decide which goals to target and learn about as a function of their current capabilities \cite{florensa2018automatic, fournier2019clic, colas2019curious, racaniere2019automated, stooke2021open}. Thus, by contrast to \irl agents, autotelic agents are {\bf open-ended} and generally {\bf few shot} learners. Note however that very few works focus on the importance of {\bf hierarchical learning} in such agents \cite{eppe2022intelligent, etcheverry2020hierarchically}.
By contrast, there is a growing tendency to combine autotelic architectures with language learning capabilities. Many papers combining goal-conditioned \rl and language understanding have recently flourished under the banner of {\em instruction following agents} \cite{luketina2019survey}, but the corresponding agents are not truly autotelic in the sense that they do not set their own goals. Autotelic agents endowed with language learning capabilities are covered in Section~\ref{sec:and}.

Thus, autotelic reinforcement learners are endowed with a lot of the properties that are missing to their interactive counterparts. But symmetrically, they generally model isolated agents, thus {\bf they miss all the interactive and social learning properties listed in Section~\ref{sec:teach}}.

\subsection{Inferential Social Learners}
\label{sec:isl}

While \irl research provides efficient ways for AI agents to learn from demonstrations, feedback or instructions, they lack methods for implementing the richest social interactions between a tutor and a tutee presented in Section~\ref{sec:teach}. A key to model these richer interactions consists in simultaneously addressing tutoring and learning processes.

A new family of agents that we call {\em inferential social learners} have started to implement these more elaborated types of interactions where the tutor and the learner build a model of each other and infer the other's beliefs, desires and intentions using these models.
These models are inspired by the inferential social learning framework of \cite{gweon2021inferential}, which considers that learners recover the meaning of underlying others' actions by inverting intuitive causal model of the way others think, plan and act. Such inference mechanisms are crucial for social learning, notably to improve the efficiency of the tutoring process. 

By drawing inspiration from language-based communication, inferential social learning goes beyond literal interpretation of actions and considers pragmatic inference \cite{grice1975logic}. The current approach of inferential social learning mechanisms exploits probabilistic inference over structured representations of the world, the other and even the representations of the other. In \cite{GOODMAN2016818}, the authors describe the rational speech act (RSA) for pragmatic reasoning in language understanding. Their approach results in probabilistic models of pragmatic speakers and listeners able to capture the meanings of complex phenomena of linguistic interaction. Taking a more developmental perspective, some other works try to account for the way parents employ ``motherese" \cite{gleitman1984current,saintgeorges2013} to talk to their children \cite{lim2014mei}, so as to endow social agents with the roots of {\bf language proficiency}. 

Transposing these linguistic considerations to the domain of interactions with objects, some works study communicative demonstrations \cite{ho2016showing}. Such demonstrations are not just directed towards the manipulated objects, but also accompanied with non-verbal cues such as eye gaze and or exaggerations of the demonstrations in the space–time dimensions used to convey the pedagogical intent. Again, similarly to the language case, some studies have focused on ``motionese,'' the non-verbal equivalent of motherese \cite{nagai2007can, nagai2009computational}. The corresponding agents can partly be seen as {\bf autotelic}, as they have some desires and intentions, but as the corresponding works focus on the interaction itself, generally  {\bf they do not come with sophisticated agents capable of open-ended learning, few shot learning or hierarchical learning}.

The mechanism behind these studies always requires to have models of the other's beliefs, desires and intentions and to reason about them to choose the most appropriate way to communicate, for instance selectively choose when to provide feedback and corrections \cite{ho2017social}.

In the Human-Robot Interaction community, similar works about agents not just intending to perform the ordinary action but also to convey something about it are using the concept of {\em legibility} \cite{dragan2013legibility, lichtenthaler2016legibility}. The key idea is to design transparency through motion models by which a robot communicates its intent to a human observer \cite{wallkotter2021explainable}. The main assumption is that humans will be able to infer the robot's intention from its motion by inverting a generative model. Thus, such works account for {\bf both forms of transparency}.

In \cite{ho2016showing}, the authors proposed a pedagogical model based on Bayesian Inference to generate communicative demonstrations. They argue that this model should help the teacher select examples to communicate a concept to the learner. They further performed experiments involving real human instructors and showed that the results were aligned with their proposed model. A more recent work proposed a demonstration-based {\bf ZPD teaching strategy} where demonstrations are not perfect, but are adapted to the current capabilities of the agent \cite{seita2019zpd}.

Driven by the \irl framework where agents learn from an external reward function, some works also study how an agent can infer information about a reward function from observed premises in the tutoring context \cite{reddy2020learning, bobu2020feature, jeon2020reward}. Moving to autotelic agents, similar processes could be transposed to infer a goal rather than a reward function. A few recent works start addressing this issue by captioning the goal of a demonstration through natural language and a goal generator \cite{zhou2020inverse, nguyen2021interactive}. We expect follow-up of such works to contribute to answering the key question of the nature of the information conveyed by the tutor through the task channel \cite{macglashan2015between, ho2016showing, ho2017social}.
 
As we noted above, social inferences require that both learners and teachers reason about the beliefs and intent of each other.
Thus, the corresponding agents need to be endowed with a Theory of Mind (\tom). There has been recent attempts to account for the acquisition of a \tom through inverse \rl \cite{jara2019theory} and in the domain of multi-agent \rl \cite{nguyen2020theory}, but these works generally suffer from the same limitations as \irl approaches: they do not consider explicit goals nor social inference processes. In robotics, efficient human-robot collaboration seems to require \tom models \cite{chakraborti2017ai}.

To summarize the whole section, \irl agents usually fall short in terms of autonomy, lacking all the properties of autotelic agents and the inferential capabilities of inferential social learners. Reciprocally, standard autotelic agents are not teachable at all. This is only when combining all approaches that future agents will display both capabilities to learn on their own and to be pedagogically taught. We now turn towards preliminary attempts in this direction.


\section{First Steps Towards Teachable Autotelic Agents}
\label{sec:and}

As outlined in the previous section, on the one hand, interactive reinforcement learners and inferential social learners are teachable, but they are not autonomous. On the other hand, autotelic agents are autonomous, but they are not teachable. To fully benefit from the efficiency of assisted discovery, agents should be simultaneously autonomous and teachable. In this section, we first describe preliminary research aiming at making autotelic agents more teachable and modelling efficient tutoring processes.

\begin{figure}[!ht]
  \centering
{\includegraphics[width=\columnwidth]{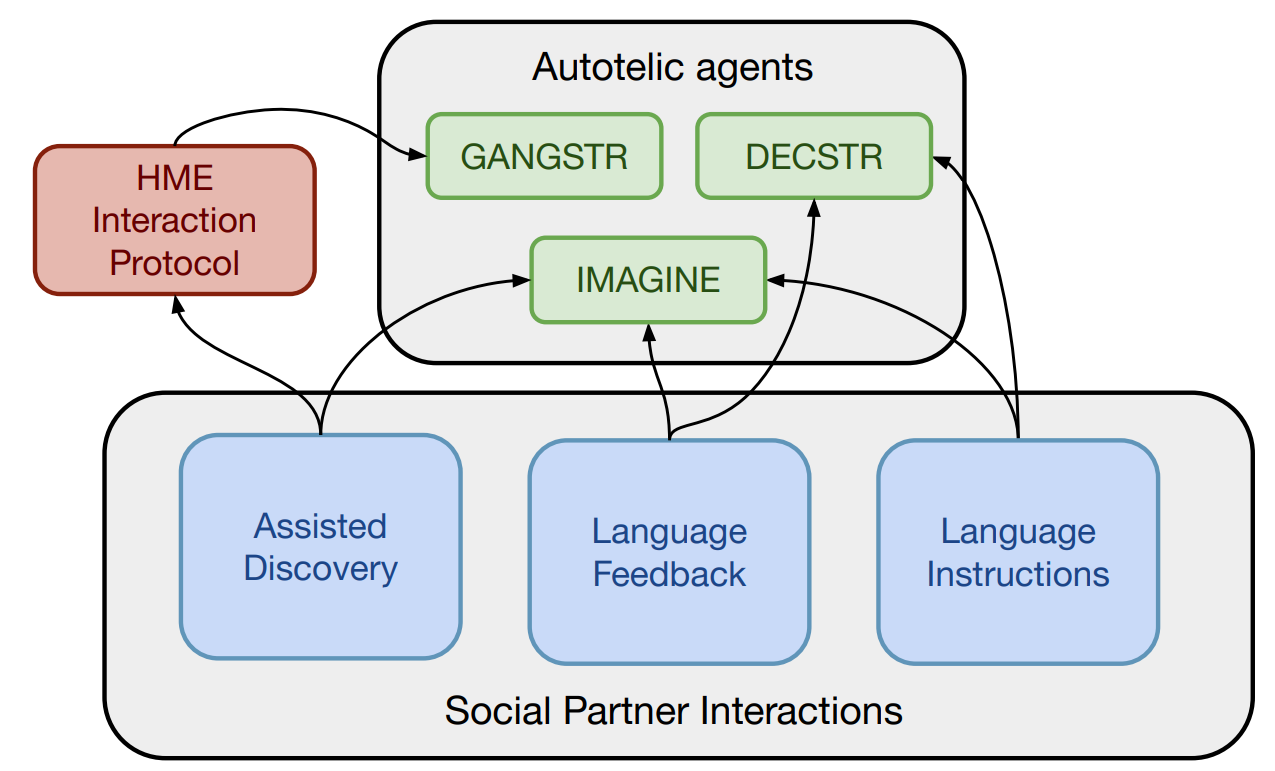}} \hfill
  \caption{Overview of the teachable autotelic agents presented in this section. \label{fig:ta_agents}}
\end{figure}

In this section we describe three autotelic agents endowed with some teachability properties, as well as a tutoring protocol, as depicted in \figurename~\ref{fig:ta_agents}. The \imagine agent learns to achieve goals described through language by a social partner, and can build on the compositionality of language to imagine new goals and pursue them. The \decstr agent represents its sensorimotor interactions with objects using abstract spatial predicates and learns to achieve goals represented in terms of these predicates. A social partner describes the performed actions and the corresponding linguistic input can be turned into instructions. The \gangstr agent extends \decstr with better sensorimotor learning capabilities and is trained with a tutoring protocol called ``Help Me Explore" (\hme) which significantly improves its capability to be taught. 

\subsection{The IMAGINE agent}

\begin{figure}[!ht]
  \centering
{\includegraphics[width=\linewidth]{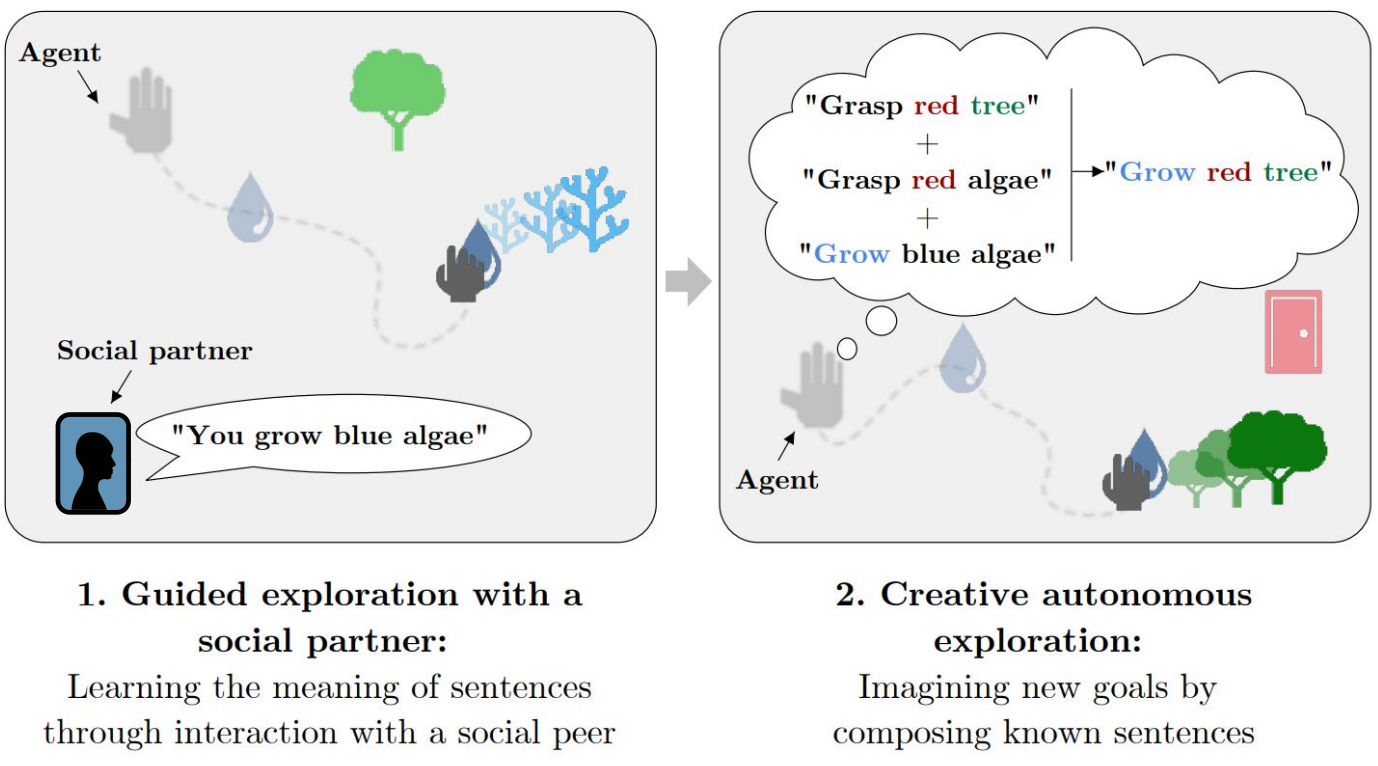}} \hfill
\vspace{-0.5cm}
   \caption{The \imagine architecture as a Vygostkian Deep RL system. 
   The agent learns to represent and understand language as a pre-existent social structure through social interactions with a synthetic tutor (left). This internalization of social language opens the door to a cognitive use of language. The agent can now imagine new goals as systematic recombinations of known sentences. As it pursues its own invented goals, \imagine creatively explores its environment (right). \imagine also leverages Vygotsky's notion of \textit{zone of proximal development} (\zpd) \cite{doolittle1997vygotsky}. In each episode, the tutor sets the scene to provide optimal challenges: it introduces the necessary objects for the agent to reach its goal (not too hard), but generates them procedurally and add distracting ones (not too easy).\label{fig:imagine}}
\end{figure}

The \imagine agent, illustrated in \figurename~\ref{fig:imagine},
is a autotelic agent, thus {\bf it benefits from all the properties of autotelic agents, apart from hierarchical learning}. The aim of \imagine is to discover and master possible interactions in a \textit{Playground} environment filled with procedurally-generated objects. As it freely explores its world by pursuing its own goals, it receives simple linguistic descriptions of interesting behaviors from a simulated tutor. It then leverages both the communicative and cognitive functions of language to benefit from these signals \cite{colas2020language}.

Let us first discuss the communicative function. If the agent hears ``you grasped a red rose,'' it turns this description into a potential goal and will try to grasp red roses again. To do so, it needs to understand what that means and to learn to replicate the interaction. The just-received description is an example of aligned data: a trajectory and a corresponding linguistic description. This data can be used to learn a goal-conditioned reward function, i.e., a function that helps the agent recognize when the current state matches the linguistic goal description. Given some examples, the agent correctly recognizes when goals are achieved and can learn a policy to perform the required interaction via standard \rl using self-generated goals and rewards. Here, \imagine uses the communicative function of language, it learns to represent the embedding of goals and goal-conditioned reward function from linguistic social interactions. As the social partner chooses to describe some interactions and not others, it effectively guides goal representation learning in the agent. Thus the goal representations of the \imagine agent are influenced by social interactions and, once they are formed, the agent can pursue them without relying on tutors.

Now, \imagine also uses a cognitive function of language. Once language has been grounded as described above, \imagine leverages the \textit{productivity} of language to generate creative goals falling outside of the domain of effects it already experienced. Language\;---\;and its compositional properties\;---\;is here used as a cognitive tool to facilitate the composition and imagination of novel goals. The mechanism is crudely inspired from usage-based linguistic theories \cite{tomasello_twenty-three-month-old_1993, tomasello_item-based_2000, goldberg_constructions_2003}. It detects recurring linguistic patterns, labels words used in similar patterns as \textit{equivalent} and uses language productively by switching equivalent words in the discovered templates. This simple mechanism generates truly creative goals that are both novel and appropriate, two adjectives used to define \textit{creativity} as discussed in \cite{runco_standard_2012}. As the authors show, this simple mechanism empowers the creative exploration of the environment and enhances the systematic generalization abilities of the agent. Whereas the communicative function of language mostly helps internalize the goal representations of social partners, the cognitive function provides the agent with individuality and open-endedness: it can generate novel creative goals that its tutor did not mention.

Using descriptions rather than instructions in \imagine is a deliberate choice given that linguistic guidance through descriptions is a key component of how parents teach language to infants \cite{yoshida_sound_2003,tomasello2009constructing}. This contrasts with the instruction-based approach dominant in \textit{language-conditioned} agents research \cite{luketina2019survey} but rarely seen in real parent-child interactions \cite{bornstein1992maternal}. Finally, the technical cornerstone of this work is that it learns a goal-conditioned reward function where the goal is learned as a language expression from data coming from the social partner. This removes the need for a lot of feedback from the social partner, which is one of the target functionalities of teachable autonomous agents. 

Despite these advanced uses of language, \imagine learns simplified pattern-based sentences and, thus, cannot be said fully {\bf language proficient}. Besides, the description of actions plays a direct role in orienting the exploration process of the agent. However, in all experiments performed in \cite{colas2020language}, the agent is only targeting its own goals and the social partner only describes what the agent does, without providing any direct incentive to target a particular goal rather than another. This is to be contrasted with the studies of \cite{wood1976role} where the tutor is in charge of making sure that children will build a pyramid. Thus, though the \imagine agent has the potential to receive instructions, this potential has not been demonstrated and \imagine cannot be said sensitive to {\bf  task-based tutoring strategy} nor to {\bf motivation regulation}. Beyond that, {\bf it does not benefit from any of the inferential social mechanisms described in Section~\ref{sec:isl}}, and it does not display {\bf social communication-based transparency}.
\subsection{The DECSTR agent}

In the intrinsically motivated agents outlined in Section~\ref{sec:pb_solvers}, just as in most goal-conditioned \rl algorithms \cite{schaul2015universal, andrychowicz2017hindsight, colas2020intrinsically}, the space of goals is generally defined as a subset of the state space of the agents \cite{florensa2018automatic,pong2019skew,nair2019contextual}. However, this approach falls short of providing the level of abstraction necessary for natural communications with social partners. The \imagine agent addresses this concern by directly representing goals in a language embedding space. But, doing so, it cannot account for the fact that infants learn to target sensorimotor goals before they master language.

The \decstr agent solves the latter issue by introducing an abstract goal representation layer in the architecture where a goal is expressed with general predicates. At the sensorimotor level, this helps targeting more abstract goals, opportunistically making profit of the current situation to obtain the easiest realisation of the goals.
The \decstr agent interacts with blocks in the {\em Fetch Manipulate Tutoring} environment, a benchmark which was used in several recent works to train hierarchical \rl agents \cite{pierrot2020learning} and autotelic agents \cite{lanier2019curiosity,colas2019curious,li2019towards}.
To take an example from this domain, having the red block ``close to" the blue block can be realized in an infinity of ways and the agent can find the simplest movement to move one of the blocks close to the other depending on the whole scene. At the language level, the abstract goal representation layer also plays a key role in grounding linguistic descriptions into sensorimotor experience, as it simplifies the correspondence between natural language instructions such as ``{\em put the red block close to the blue block}'' and the goals the agent manipulates in practice. The capability of generating a diversity of goals for a same description also results in an increased diversity of behaviors displayed by the agent and a capability to retry to pursue the same goal in another way \cite{akakzia2021grounding}.

\begin{figure}[!ht]
  \centering
{\includegraphics[width=\linewidth]{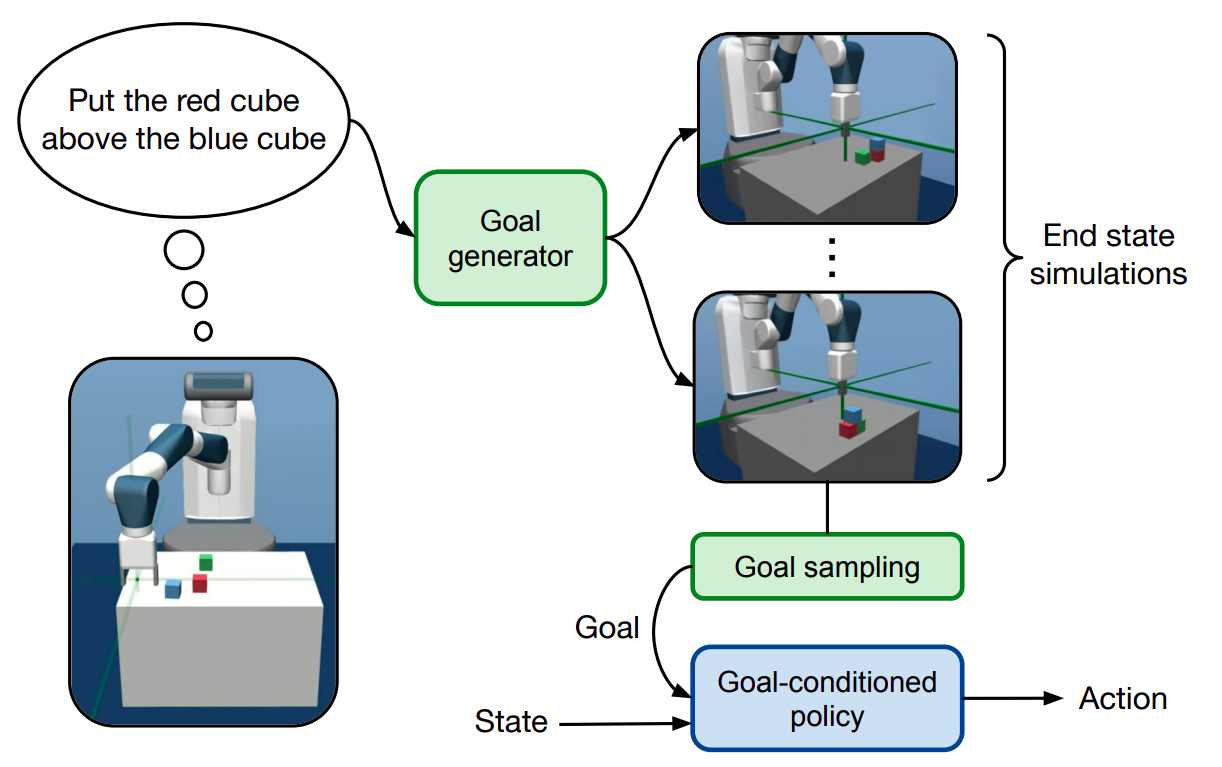}} \hfill
\vspace{-0.5cm}
   \caption{
   The \decstr architecture as a Vygotskian autotelic agent. \decstr learns to ground linguistic descriptions of its trajectories provided by a synthetic tutor into innate semantic representations. This language grounding process occurs in a \textit{language-conditioned goal generator}. Once trained, \decstr uses language as a cognitive tool to guide the simulation of possible future world configurations matching an input description (e.g.\,from the tutor or self-generated). As it selects one of them as goal, \decstr commits to turning the world into this selected future. \label{fig:lgb}}
\end{figure}

\begin{figure}[!ht]
  \centering
{\includegraphics[width=\linewidth]{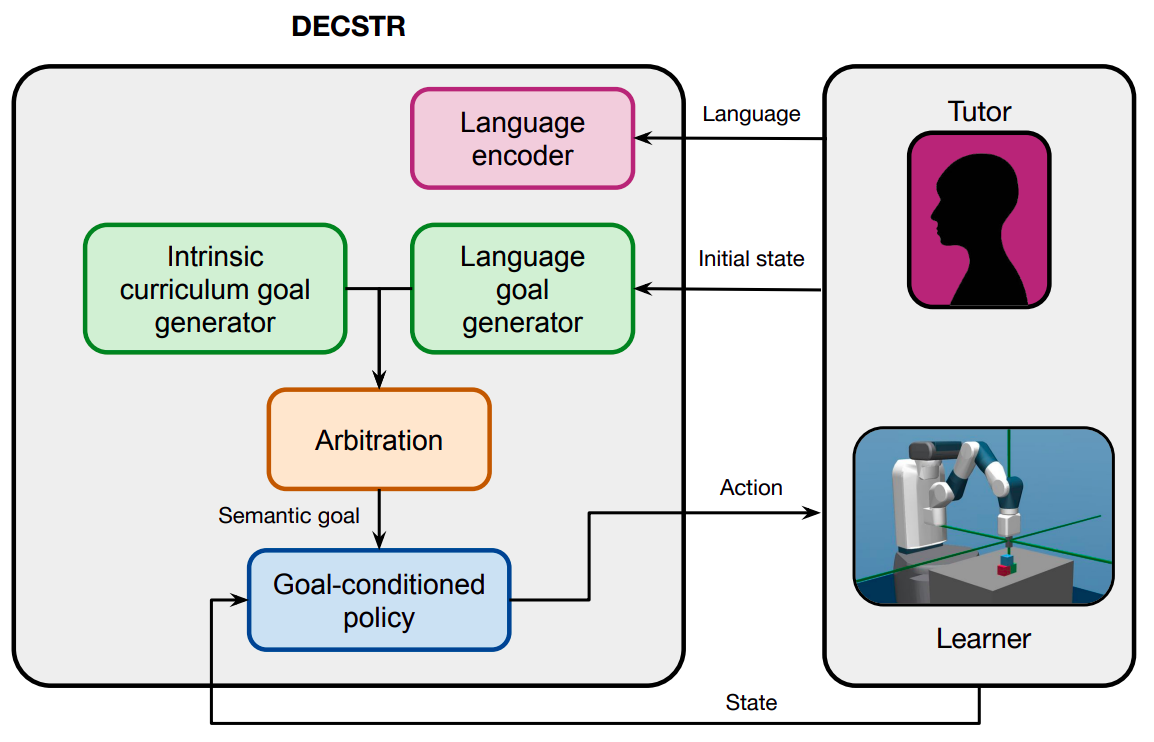}} \hfill
   \caption{
    \textit{The \decstr architecture in the Fetch Manipulate Tutoring environment}. The tutor can interact with the agent by setting the initial scene, providing descriptions or instructions. \decstr uses aligned descriptions and trajectories to train a language-conditioned goal generator mapping language to a set of matching configurations used as goals. The agent either pursues language-conditioned goals or its own. In either case, it learns to achieve them through intrinsically motivated goal-conditioned reinforcement learning.\label{fig:fetch_env}}
\end{figure}

In more details, the architecture of the \decstr agent is depicted in \figurename~\ref{fig:fetch_env}. To ground language into its sensorimotor experience, it relies on a {\em Language Goal Generator} (LGG) that takes a language expression as input and samples concrete goals matching the linguistic expression. \figurename~\ref{fig:lgb} depicts the way this component endows sensorimotor autotelic agents with language sensitivity by relating language to behavior. This is precisely through this LGG that language is grounded into sensorimotor goals.

After a while, the \decstr agent consistently learns to build pyramids and towers of three blocks. Once becoming more expert, it gets opportunistic, making profit of the current configuration to reach its goal with as few block moves as possible. More details about this work are presented in \cite{akakzia2021grounding}.

The \decstr agent {\bf benefits from nearly the same properties as  \imagine}. The main difference is that, in \decstr, since language acquisition and instructions are decoupled from sensorimotor learning, language obviously plays no role in the sensorimotor learning phase thus {\bf there is no room for social-based tutoring strategy} with \decstr.

\subsection{The GANGSTR + HME agent}

The last work we investigate combines an autotelic agent and an interaction protocol.

\vspace{3mm}
\par\noindent {\bf The GANGSTR agent}

Similarly to \decstr, \gangstr is a predicate-based autotelic agent designed for object manipulation. It faces five blocks within the Fetch Manipulate Tutoring environment and strives to discover a large spectrum of diverse semantic configurations from the easy ones (no stacks at all, stacks of two, pyramids of three) to the more complex ones (stacks of three and higher, combinations of stacks and pyramids). As opposed to the \imagine and \decstr agents, {\bf it does not rely on language as a communication tool}. However, it can decompose its goals into sequences of sub-goals. This endows it with a {\bf hierarchical learning} capability. Results show that this decomposition is necessary for the mastery of the most complex goals such as stacks of four or five blocks.

The \gangstr agent builds a structured representation of its discovered semantic goal space as a graph where nodes represent configurations. Two nodes are linked with a directed edge if it is possible to reach one configuration from the other by moving exactly one object.
In turn, this graph is exploited in the context of the \hme protocol.

\vspace{3mm}
\par\noindent {\bf The HME protocol}

Help Me Explore (\hme) is a simple interaction protocol where a tutor helps an autotelic agent explore its goal space. The role of the tutor is to propose \textit{communicated goals} that help the agent quickly discover its entire goal space. The autotelic agent then remembers the communicated goals and trains to pursue them as well as its own goals. When \gangstr uses the \hme protocol, both communicated goals and autotelic goals are subsets of the same predicate-based goal space as in \decstr.

To select a goal to communicate, the tutor has a model of the agent's current exploration limits. In \gangstr, this model is a copy of the graph of the semantic goals discovered by the learner. Using this model, the tutor first communicates a goal at the frontier of the learner's knowledge. If the learner succeeds in reaching this goal, the tutor uses its own, full-grown graph of goals to select a goal just beyond that frontier that the learner may reach with just one block manipulation from the current configuration. The learner then memorizes this (frontier, beyond) pair of goals and may train reaching them on its own until it sufficient masters it. 
In practice, the agent first discovers easy goals. Once these goals are discovered, the agent can either follow 1) autotelic episodes by uniformly sampling a goal to pursue among the set of autonomously discovered goals, or 2) social episodes where the tutor communicates a pair of goals as explained above.

The \hme protocol accounts for several features of human tutoring processes described in Section~\ref{sec:prop_tutor}. First, memorizing the (frontier, beyond) pair of goals is a form of internalization mechanism where the learner can train on its own on goals communicated by the tutor. Moreover, by proposing (frontier, beyond) pair of goals, {\bf the tutor maintains the learner into its \zpd}, where goals at the frontier act as stepping stones towards further discoveries.
Even more strikingly, an experimental study where the rate of social episodes is varied from 0\% (pure autotelic learning) to 100\% (pure instruction following) shows that a moderate rate of social episodes works best. This is clearly reminiscent of the claim in educational sciences that guided play and assisted discovery work better than purely autonomous learning or direct instructions.
More details about \gangstr and the \hme protocol are presented in \cite{akakzia2022help}.

\vspace{3mm}
\par\noindent {\bf Properties of GANGSTR + HME}

The properties of \gangstr + \hme are close to those of the \decstr agent. We already outlined that it benefits from {\bf hierarchical learning} but {\bf not from language proficiency}. Additionally, by contrast with \decstr, \gangstr incrementally learns to reach goals in interaction with its tutor, who can thus play a much more direct role in orienting the exploration process of the agent. Thus \gangstr + \hme can thus be said sensitive to {\bf motivation regulation} and to {\bf  task-based tutoring strategy}.

As for missing properties, \gangstr + \hme, when the tutor suggests a goal, the agent immediately pursues it. A more realistic agent should be endowed with mechanisms to arbitrate between their own goals and those coming from the tutor. These arbitration mechanisms should, in turn, endow these agents with a form of `personality' where different agents with different parameters would be more or less teachable, exactly as children. But it would also require a form a {\bf social communication-based transparency} which is currently missing in the autotelic agents we have listed.

\section{Discussion and Future Directions}
\label{sec:discu}

For all the properties we have listed in Section~\ref{sec:teach}, Table~\ref{tab:teachability_checklist} recaps whether all the agents that we have presented so far display or not these properties.

One can see that \imagine, \decstr and \hme + \gangstr all display some of the properties one may expect from teachable autotelic agents. They are even complementary with respect to several of these properties. For instance, \imagine and \decstr are endowed with a basic capability to interpret language, which is not the case of \hme + \gangstr. But the latter implements reciprocal modelling of the tutor and the learner, by contrast with the former methods. Thus a good deal of the limitations of the above agents could be overcome by combining their capabilities. 

Integrating all these features into a single teachable autonomous agent would significantly increase their flexibility, making it possible to leverage a more significant part of the vast repertoire of interaction protocols or ``{\em pragmatic frames}'' used in human tutoring \cite{vollmer2016pragmatic} and, more generally, opening the possibility to more natural, non template-based interaction \cite{zhou2020inverse}. It would also open the way towards new research questions, such as modelling feedback about goal selection rather than just about action, or the need for an arbitration mechanism between intrinsic motivations and social feedback. Besides, such an integration may require some developmental dynamics, resulting in displaying {\em overlapping waves} of sensorimotor and linguistic development \cite{siegler1998emerging}. 

However, Table~\ref{tab:teachability_checklist} shows that such a combination would still lack most of the properties of inferential social agents. Thus an important direction for future research will consist in combining the properties of the existing teachable autotelic agents and endow them with further social inference capabilities. To do so, the model of the learner's skills used in the \hme protocol could serve to leverage the Bayesian inference processes proposed in \cite{velez2021learning}. Besides, all tutoring processes that \imagine, \decstr and \hme + \gangstr agents can benefit from consist in descriptions and instructions about goals. These agents could be extended with capabilities to learn from pedagogical demonstrations.

Looking more closely at Table~\ref{tab:teachability_checklist}, one can see that an agent combining the properties of \imagine, \decstr, \hme + \gangstr and inferential social learning agents would still lack one property, which is language proficiency. In language-augmented agents such as \imagine and \decstr, the template-based language learning mechanism does not mimic the language acquisition processes of infants. Though, the developmental processes and constraints involved in language acquisition may play a crucial role in the more general acquisition of interaction capabilities. 
Thus, more realistic models of language acquisition and learning of semantic predicates in infants \cite{goldstein2003social, kuhl2004early, mandler2012spatial} may be required to better account for these capabilities. Besides, communication itself could be extended to richer natural language interactions \cite{lynch2020grounding} so as to increase the teachability of these agents. The very fast progress in language learning with large transformer models \cite{floridi2020gpt,rae2021scaling} creates an opportunity in this direction.

Finally, in contrast with the above questions which should be answered soon given the current pace of learning agents research, there are a few questions which remain largely unaddressed. For instance, how can an autonomous agent learn to determine the positive or negative valence of sophisticated feedback signals such as attitudes or linguistic nuances? Or how can we endow agents with the capability to generalize immediately what they have learned from an intended demonstration to other contexts? More fundamentally, how can we extend the {\em language grounding} capabilities of current teachable autotelic agents to solve the harder {\em symbol grounding} problem \cite{harnad1990symbol}? In short, language in \decstr and \imagine is more indexical than symbolic because the language tokens do not form a system \cite{nieder2009prefrontal}. The acquisition of symbolic behavior is an emerging topic \cite{santoro2021symbolic} which can be of fundamental importance for future agents as, from one side, considering a social partner is necessary to establish conventional meaning and, from the other side, such agents may need the flexibility of symbolic behavior to appropriately learn from natural tutors.

\section{Conclusion}
\label{sec:conclu}

Many efforts have been made to endow artificial agents with the capacity to learn from humans, in a natural and unconstrained manner. However, for now, we are still far from achieving ``normal teaching of a child,'' in reference to Turing's view. 
In this paper, by investigating the way children are taught, we claimed that autotelic agents were a better starting point for such a research than standard RL agents. The resulting agents would pursue their own goals, but should be endowed with an additional capability to be taught so that they choose their goals in accordance with the expectation of their users. We have then described some of the ongoing and immediate future work along this line of research, and revealed some of the issues which must be overcome to get closer to the way children are taught.

In the immediate future, the existing teachable autotelic agents could be combined and integrated with inferential social learning capabilities and more natural language learning capabilities. Given the fast progress currently observed in the design of autotelic learning agents, we expect to soon see good enough teachable autonomous agents to use them for quantitative analyses in developmental psychology studies and for a better design of education programs. We also believe that this starting point is a key move towards better insertion of AI agents in the society, with improved capabilities to communicate with and to adapt to their human users, which is one of the central concerns of AI research.

\section*{Acknowledgments}

The authors would like to thank Katarina Begus for advice about an early version of this paper.

This  project  has  received  funding from European Union’s  Horizon 2020 ICT-48 research and innovation actions under grant agreement No 952026 (HumanE-AI-Net) and from  the  European Union’s  Horizon 2020 research  and  innovation programme under grant agreement No 765955 (ANIMATAS).

\bibliographystyle{IEEEtran}
\bibliography{ttaa}

\end{document}